\pdfoutput=1

\documentclass[11pt]{article}

\usepackage{acl}

\usepackage{times}
\usepackage{latexsym}
\usepackage{booktabs}

\usepackage[T1]{fontenc}

\usepackage[utf8]{inputenc}

\usepackage{microtype}

\usepackage{array}
\usepackage{url}
\usepackage{booktabs}
\usepackage{algorithm}
\usepackage{algpseudocode}
\usepackage{arydshln}
\usepackage{graphicx}
\usepackage[shortlabels]{enumitem}

\newcolumntype{P}[1]{>{\centering\arraybackslash}p{#1}}

\title{Graph-based Keyword Planning for Legal Clause Generation from Topics}

\author{\textup{Sagar Joshi*}, \textup{Sumanth Balaji*} \\
  IIIT Hyderabad \\
  \texttt{\{sagar.joshi,sumanth.balaji\}@research.iiit.ac.in} \\\AND
  Aparna Garimella \\
  Adobe Research \\
  \texttt{garimell@adobe.com} \\\And
  Vasudeva Varma \\
  IIIT Hyderabad \\
  \texttt{vv@iiit.ac.in}
  }

\begin{document}

\maketitle
\def\thefootnote{*}\footnotetext{Both the authors have equal contribution to this work.}
\def\thefootnote{\arabic{footnote}}

\begin{abstract}
Generating domain-specific content such as legal clauses based on minimal user-provided information can be of significant benefit in automating legal contract generation.
In this paper, we propose a controllable graph-based mechanism that can generate legal clauses using only the topic or type of the legal clauses.
Our pipeline consists of two stages involving a graph-based planner followed by a clause generator.
The planner outlines the content of a legal clause as a sequence of keywords in the order of generic to more specific clause information based on the input topic using a controllable graph-based mechanism.
The generation stage takes in a given plan and generates a clause.
The pipeline consists of a graph-based planner followed by text generation.
We illustrate the effectiveness of our proposed two-stage approach on a broad set of clause topics in contracts.
\end{abstract}

\section{Introduction}

\begin{figure*}
    \centering
    \includegraphics[scale=0.425]{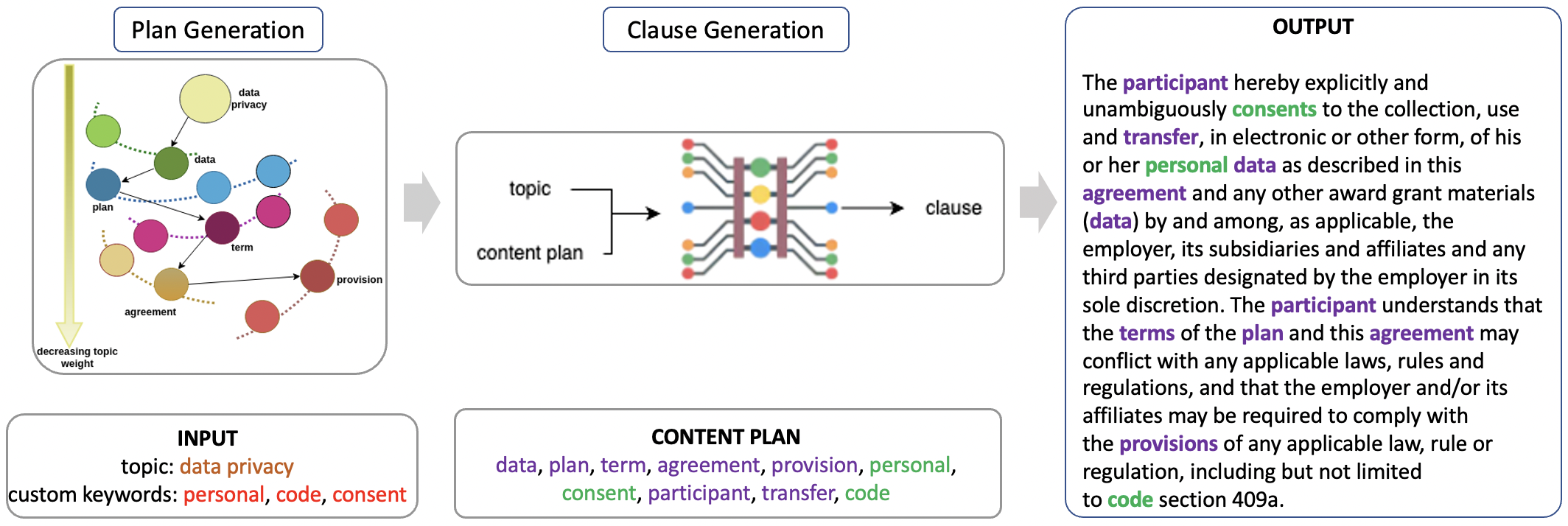}
    \caption{The proposed 2-stage clause generation pipeline: User specifies the input topic (\textit{data privacy}) along with a set of keywords for customization (\textit{personal, code, consent}). The first stage involves generating a customized content plan, including the custom keywords (green) along with other plan keywords derived purely from a graph walk (purple). In the second stage, a clause generator interpolates through the keywords in the plan to produce a coherent, meaningful legal clause for the given topic.}
    \label{fig:intro}
    \vspace{-0.15in}
\end{figure*}

Contracts are essential discourse units that frequent in several day-to-day business workflows, especially between companies and governmental organizations.
The fundamental units of discourse in contracts consist of ``clauses'' that are paragraphs of text that outline the terms and conditions of various types or {\it topics} (e.g., {\it severability}, {\it benefits}) (Table \ref{tab:sample_clause}).
Legal clauses can be characterized by their high inter-sentence similarity, and topic-specific content \cite{Simonson2019TheEO}.
For example, \citet{NLP-and-legal} showed that the sentences in legal corpora are almost $20\%$ similar to each other.
Drafting contracts by legal counsel is a manual process of taking a skeletal set of clauses and adding or modifying them for the contract goal.
Given their highly domain-specific content and unique linguistic structure, contract drafters in legal counsel can significantly benefit from applying Natural Language Processing (NLP) techniques to aid contract creation \cite{NLP-and-legal}.

\begin{table}[t]
\centering
\scriptsize
\begin{tabular}{p{7.2cm}}
\toprule
In case any provision herein or obligation hereunder or any Note or other Credit Document shall be invalid, illegal, or unenforceable in any jurisdiction, the validity, legality and enforceability of the remaining provisions or obligations, or of such provision or obligation in any other jurisdiction, shall not in any way be affected or impaired thereby. \\
\bottomrule
\end{tabular}
\caption{An example \textit{severability} clause from a legal contract.}
\label{tab:sample_clause}
\vspace{-0.15in}
\end{table}

There have been recent advances in Transformer-based \cite{10.5555/3295222.3295349} methods for text generation in varied flavors, such as prompt-based causal generation \cite{radford2019language}, conditional generation based on control codes \cite{CTRL}, and retrieval-augmented generation based on queries \cite{10.5555/3495724.3496517}.
However, these methods are primarily studied in generic NLP domains, and legal text generation remains largely unexplored. 
The only previous work that addressed the task of legal text generation is {\sc ClauseRec} \cite{aggarwal-etal-2021-clauserec}, in which a Transformer-based decoder is trained to generate missing legal clauses in a given contract document, conditioned on the clause topic and the content in the contract.
However, \citet{aggarwal-etal-2021-clauserec} noted that the clauses generated by {\sc ClauseRec} suffer from low linguistic variations within topics, thus resulting in content that is thematically relevant but missing a few nuances.
In general, we believe conditioning text generation on only the high-level clause topics or the contract content may not capture the subtleties present in legal clauses, hence call for an iterative approach to learn the clause-specific content in a top-down manner.

We find inspiration in the content planning paradigm for story generation \cite{mcintyre-lapata-2010-plot,plan_and_write,Graphplan} in which an intermediate plan (e.g., a set of keywords) is used to generate final stories. 
In this paper, we study legal text generation, and propose a two-staged pipeline (Figure \ref{fig:intro}) to generate legal clauses from topics iteratively using keywords. 
Specifically, the first module of our pipeline comprises of a graph-based planner that takes in a topic (and an optional set of keywords) and produces an ordered content plan consisting of keywords, with those more generic to the topic ranked higher, and those more specific to a clause ranked lower.
Note that unlike \cite{aggarwal-etal-2021-clauserec}, which aimed to generate missing clauses given a contract, we focus on generating legal clauses only based on a given topic and an optional set of keywords.
We approximate the generated content plan to the user intent, which will be translated into a legal clause.
In the second stage, the plan is used as a control mechanism, and a Transformer-based language model is trained to generate legal clauses conditioned on the topic and the plan.\footnote{The code for this work can be found here: \url{https://github.com/sagarsj42/legal_clause_gen_from_topic_keywords}}

Following are the main contributions of this paper:
{\bf (1)} We propose a novel two-staged pipeline for legal clause generation, comprising a content planner that generates a keyword-based plan, and a content generator that generates legal clauses conditioned on the clause topics and keyword plans.
Our proposed content planner consists of a simple, lightweight graph-based mechanism that performs a graph walk using the input topic to generate a plan consisting of generic to specific keywords. The plan can be customized by specifying a few control keywords, bringing controllability to the generation.
{\bf (2)} We compare our approach with several conditional and causal text generation baselines, and illustrate strong empirical results for legal clause generation.
{\bf (3)} We also show that our approach can be generalized well to a diverse range of clause topics, thus indicating the extensibility of our approach for legal text generation.
We believe our work takes one step further in the area of automatic clause generation for AI-aided drafting of legal documents.

\section{Graph-based Planning for Clause Generation}

Our proposed approach aims to generate legal clauses to aid legal counsel in contract drafting.
To do so, it takes as input a clause topic (e.g., {\it data privacy}) along with a few keywords for customization ({\it personal}, {\it code}, {\it consent}), and generates more keywords in order to obtain a customized content plan as illustrated in Figure \ref{fig:intro}.
This keyword-based plan, which we consider an approximation to user-specified keywords per their preferences, is then used to generate a meaningful legal clause.

\subsection{Dataset creation}

\noindent\textbf{Ranked keyword extraction per topic.}
For every clause topic $t$ from the set of all clause topics $T$, we extract an ordered set of keywords $K^t = \{k_1^t, k_2^t, …., k_{m_1}^t\}$ representing the topic using an off-the-shelf keyword extractor.
Each extracted keyword is a single, comprehensible word unit occurring in a clause.
The ordered set of keywords under a topic represents the salient words under that topic, approximately ranked based on their prominence.
In the ranked order, the words more generic to the topic (perceived to carry more information about the topic's generic form\footnote{By topic's generic form, we refer to the clause content that most commonly occurs across clauses under that topic, being characteristic to that topic.}) are ranked higher. The ones less generic to the topic but more specific to individual clauses (perceived to be more characteristic of an individual clause) are ranked lower.
The keywords are lemmatized using a WordNet-based lemmatizer.
\begin{figure}[t]
    \centering
    \includegraphics[scale=0.4]{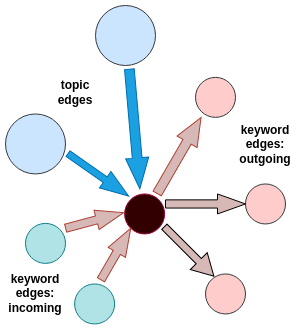}
    \caption{View of the types of node connections in the directed graph $G$.}
    \label{fig:graph_node}
    \vspace{-0.15in}
\end{figure}

\noindent\textbf{Reference keyword plans for clauses.}
Each clause is represented by a reference keyword plan which consists of a ranked list of keywords corresponding to the topic of that clause.
For every clause in a topic, we check for the existence of each ranked topic keyword and sequentially add them as plan keywords, thus preparing a ranked plan of keywords appearing in the clause.
Applying for all the topics, we have a dataset of clause-keyword plans $D^t = (C^t, R^t)$ for each topic $t$ in the dataset. $C^t$ represents the set of clauses $\{c_1^t, c_2^t, ..., c_{m_2}^t\}$ under the topic, and $R^t$ represents the set of corresponding reference keyword plans for the clauses, with the plan for a clause $c$ being a ranked list of keywords, $r^c = [r_i^c]_{i=1}^{i=n}$ (where $r_i^c$ represents keyword selected for each stage $i$). 
\subsection{Graph construction}


A single, directed graph $G$ is constructed to capture the keyword plan information from all the topics in a unified representation as illustrated in Algorithm \ref{alg:gc}.
The graph $G$ is initialized with the set of nodes $N = T \cup K$ consisting of all the topics $T$ and an accumulated set of keywords from the topics, $K = \bigcup_{t \in T} K^t$.
Each node in the graph has incoming connections from relevant topic nodes along with incoming and outgoing connections to keyword nodes, as shown in Figure \ref{fig:graph_node}.

\begin{algorithm}[t]
\scriptsize
\caption{Graph construction}
\label{alg:gc}
\begin{algorithmic}[1]
\Require{Topics $T$, keywords $K$, reference plans $R = \bigcup_{t \in T} R^t$}
\State{Initialize graph $G$ with nodes $N \gets T \cup K$}
\State{Edges $e(n_1, n_2) \gets 0 \;\; \forall (n_1, n_2) \in N$}
\For{topic $t \in T$}
\State{Topic frequency, $f \gets len(R^t)$}
\For{reference plan $r^t \in R^t$}
\For{step $s \gets 1 \; to \; n$}
\State{Step value, $v = 1/(s \cdot f)$}
\State{$e(t, r_s^t) \gets e(t, r_s^t) + v$}
\State{$e(r_{s-1}^t, r_s^t) \gets e(r_{s-1}^t, r_s^t) + v$}
\EndFor
\EndFor
\EndFor
\State\Return{$G$}
\end{algorithmic}
\end{algorithm}
Edges weights between every pair of topic-keyword and keyword-keyword nodes are calculated based on their occurrence in the train set as demonstrated in the algorithm.
In this process, we walk through the reference plan for each clause under all the topics in the train set.
A topic-keyword edge $(t, r_s^t)$ is added for the occurrence of a keyword in a reference plan as a stage $s$ of the plan.
Similarly, a keyword-keyword edge $e(r_{s-1}^t, r_s^t)$ is added for the occurrence of consecutively occurring keywords $r_{s-1}$ and $r_s^t$ in the plan.
Every occurrence of such topic-keyword or keyword-keyword pair adds an incremental weight to the corresponding edge.
The weight given to each occurrence $v$ is normalized by (1) no. of clauses $f$ under that topic and (2) the stage value $s$ at which the occurrence occurs.
(1) accounts for the substantial imbalance in the clause type frequencies for the topics in the dataset, while (2) gives lesser importance to the keywords present at lower stages of the plan, thus statistically recording the generic to specific order within the edges of the graph.

\subsection{Plan generation}

\begin{figure}[t]
    \centering
    \includegraphics[scale=0.8]{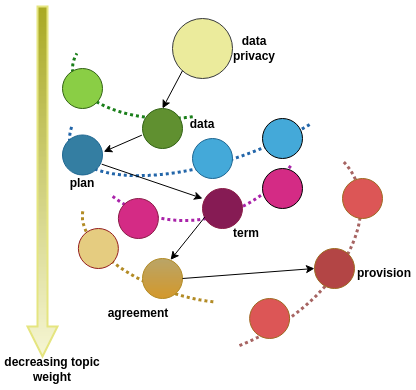}
    \caption{Illustration of the first 5 stages of freeform plan generation, i.e., without using custom keywords for control. Given input topic: \textit{data privacy}.}
    \label{fig:plangen}
    \vspace{-0.15in}
\end{figure}

\begin{algorithm}[h]
\scriptsize
\caption{Plan generation}
\label{alg:pg}
\begin{algorithmic}[1]
\Require{Graph $G$, topic $t$, stepwise thresholds $TH$, optional custom keywords $q_c$}
\State{\textbf{Initialize:} Plan $q \gets []$; Current node, $cn \gets t$}
\For{step $s \gets 1 \; to \; n$}
\State{Neighbors of current node, $N \gets Neighbors(G, cn)$}
\State{\# candidates to score, $l \gets len(N)$}
\State{Neighbor scores, $S \gets Zeros(l)$}
\For{$i \gets 1 \; to \; l$}
\State{$cand \gets Neighbors[i]$}
\State{$S[i] \gets G.edge\_score(t, cand) + G.edge\_score(cn, cand)$}
\EndFor
\State{$SORT(N)$ by $S$, in descending order}
\State{Top candidates, $TC \gets N[:TH[i]]$}
\If{$\exists k \in q_c$ s.t. $k \in TC$}
\State{$cn \gets k$}
\State{$q_c \gets q_c - k$}
\Else
\State{$cn \gets GET\_RANDOM(TC)$}
\EndIf
\State{$APPEND(q, cn)$}
\EndFor
\State{\Return{$q$}}
\end{algorithmic}
\end{algorithm}

\noindent\textbf{Plan generation.} Once we have the graph $G$, a plan $q$ is generated at inference time by walking on the graph using the given input topic $t$ as the starting point as shown in Algorithm \ref{alg:pg}.
The provided input can also contain additional keywords $q_c$ to be included in the plan, based on which the generated plan $q$ can be customized.

A walk down the graph starts from the topic node $t$ while selecting a keyword $k$ from the best neighbors of each node.
The selected neighbor then acts as the node for the next stage from which subsequent selection is to be made.
This proceeds till we complete $n$ stages of plan generation. For an appropriate selection, all neighbors of the current node $cn$ are scored and ranked before making a selection.
The window size for selecting a neighbor from the top-ranked ones at each stage is specified by the thresholds $TH$.
For ranking the neighbors, each candidate $cand$ is scored based on the sum of their edge scores to the topic $G.edge\_score(t, cand)$ and the current node $G.edge\_score(cn, cand)$, thus facilitating the selection of keywords relevant to the topic and the current node context.
At every stage, if the list of top-ranked candidates contains one of the custom keywords $q_c$, we directly select the custom keyword and remove it from $q_c$ to avoid further repetition of that word in the plan.
The generated plan can thus be given as:
\begin{equation}
    q = [q_s]_{s=1}^{s=n}
\end{equation}

\subsection{Clause generation}

\noindent\textbf{Model training.}
We train a language model $LM(\theta)$ for generating a clause $c$ by conditioning on a reference plan $r^c$ and topic $t$, where $\theta$ are the parameters of $LM$.
In this, we minimize the negative log-likelihood of the probability of $c$ as given by the model:
\begin{equation}
    L_{gen} = \sum_{t \, \in \, T} \sum_{c \, \in \, C^t} -log[ \, p(c \: | \: r^c, t, \theta) \, ]
\end{equation}
This trained model can be used for generating a clause from a custom-generated plan $q$.
Since the model has seen a large number of reference plans and their corresponding clauses, the model is expected to generate the right $c$ for a $q$ given by the planner. \\

\noindent\textbf{Inferencing clauses from custom plans.}
At inference, we use the constructed graph $G$ and the language model $LM(\theta)$.
We expect a minimal input $t$ indicating the topic of the clause to be generated along with an optional set of keywords $q_{k_c}$ for customization.
We run the plan generation algorithm based on this information to obtain a custom plan $q$ as demonstrated in Algorithm \ref{alg:pg}.
The customizability in plan generation can be exploited through an iterative plan-and-generate process involving iterative modifications to the plan before achieving a user-desired state of the clause.
Appendix \ref{appdx:iterative-flow} illustrates such an example flow of plan modification followed by subsequent clause generation to enable an end user to achieve the clause in desired state.

\section{Experiments}

\subsection{Dataset}

We use the LEDGAR \cite{tuggener-etal-2020-ledgar} dataset for our experiments.
The cleaned version of this dataset consists of 60,540 contracts extracted from the EDGAR \cite{loukas-etal-2021-edgar} database containing 846,274 clauses (or ``provisions'') from 12,608 topics (or ``labels'').
We first create splits of the dataset at the contract level to ensure no data leakage in evaluation by making train, dev, and test sets made in a proportion of 85:5:10.
We discard those clauses which belong to more than one topic in the subsequent experiments.
From the train set contracts, we select those clause topics with a minimum clause frequency of 100, resulting in 387,210 clauses from 939 topics for training.
We use these selected topics for identifying applicable clauses in the dev and test splits.

\subsection{Keyword extraction and graph construction}

We use the YAKE \cite{yake} keyword extractor for extracting keywords.
Using YAKE allows us to extract a ranked order of keywords based on their prominence.
The quality of ranked keywords given by the simpler statistical algorithm in YAKE was found to align with the notion of generic to specific information flow.
To approximate the generic to specific order of keywords, we concatenate all the clauses under a topic and extract up to 200 ($m_1$) keywords per topic ($K^t$).
These extracted keywords in ranked order represent each clause ($c_i^t$) as a reference plan of keywords ($r^{c_i}$) in which we limit the number of keywords per clause to 10 ($n$).
The dataset of clause-plans ($\bigcup_{t \in T}D^t$) thus obtained is used to construct the graph ($G$) which consists of 267,893 edges ($e(. \:,\:.)$) and 46,953 nodes ($N$) - with a sparsity of 99.99\% enabling it to be used as a lightweight mechanism for control.

\subsection{Experimental settings}
We experiment with pretrained GPT-2 \cite{radford2019language} and BART \cite{lewis-etal-2020-bart} models for clause generation. Both models are trained on a batch size of 32 for 15 epochs.
The learning rate schedule follows an initial warm-up till $1e-05$ for $1/4th$ of the total training steps, followed by linear decay.
AdamW \cite{Loshchilov2019DecoupledWD} optimization with a weight decay of 0.01 is used.
The maximum generation length of the clause is kept to 700 tokens.

GPT-2 is trained in the usual causal generation paradigm in which we supply the topic concatenated with the plan as the prompt based on which the model generates a clause.
For BART, conditional generation is employed in which a topic-plan concatenated input is supplied to the encoder, conditioning on which the model is trained to output a relevant clause.

\subsection{Baselines}
\label{sec:baselines}

We consider the following baselines to evaluate the effectiveness of our plan-based approach for clause generation.
\vspace{-0.1em}
\begin{itemize}[topsep=0.1em]
    \itemsep 0em
    \item \textbf{Prompt2Clause:}
    We consider the first 10 words of a clause as the prompt (plan) and finetune a GPT-2 model for clause generation following the usual causal generation paradigm.
    \vspace{-1.5em}
    \item \textbf{Top2Clause:}
    We train a BART model to generate a clause solely conditioned on the topic of the clause.
    \vspace{-0.4em}
    \item \textbf{RandKwd2Clause:}
    Keyword order is randomized in the plan, and supplied with the topic for BART-based conditional generation.
    \vspace{-1.5em}
    \item \textbf{Plan2Clause-Retrieval:}
    We use the reference plans to retrieve from a TF-IDF-based index of clauses in the train dataset.
\end{itemize}

\section{Results}

\subsection{Plan generation}

\begin{table}[h]
\scriptsize
    \centering
    \begin{tabular}{|c|c|c|}
        \hline
        & mean & median \\
        \hline
        rank & 26.70 & 9.5 \\
        \hline
        \# neighbors & 385.62 & 327 \\
        \hline
    \end{tabular}
    \caption{Ranking generated plans based on references for estimation of plan quality generation.}
    \label{tab:plan_rank_res}
\end{table}

To estimate the generated plans' quality, we walk the graph as shown in Algorithm \ref{alg:pg}. However, we use a reference plan to determine the rank given to the expected keyword at every stage of the plan.
We take aggregated mean and median values of the ranks given at each stage across all the plans and compare them against the corresponding number of neighbor nodes encountered for ranking.
As seen in Table \ref{tab:plan_rank_res}, the graph walk gave a median rank of 9.5 against 327 neighbors.
This shows the effectiveness of a simple, lightweight graph-based modeling for generating clause plans.

\subsection{Clause generation}

\begin{table}[h]
    \centering
    \scalebox{0.7}{
    \begin{tabular}{c |c c c c}
    \hline
        Experiment & BLEU & R-1 & R-2 & R-L \\
        \hline
        Prompt2Clause & 20.2 & 19.68 & 12.56 & 16.1 \\
        Top2Clause & 33.38 & 43.32 & 24.14 & 33.74 \\
        RandKwds2Clause & 28.4 & 51.18 & 32.11 & 40.74 \\
        Plan2Clause-Retrieval & 40.74 & 48.34 & 29.57 & 38.73 \\
        Plan2Clause-GPT2 & 39.18 & 48.24 & 29.73 & 39.25 \\
        Plan2Clause-BART & \textbf{48.98} & \textbf{58.99} & \textbf{37.95} & \textbf{46.11} \\
    \hline
    \end{tabular}
    }
    \caption{Results of clause generation from plans.}
    \label{tab:clsgn_res}
\end{table}

We compare the baselines outlined in Section \ref{sec:baselines} against results based on our finetuned GPT-2 (Plan2Clause-GPT2) and BART (Plan2Clause-BART) models in Table \ref{tab:clsgn_res}.
We use BLEU \cite{papineni-etal-2002-bleu} and ROUGE \cite{lin-2004-rouge} metrics to evaluate the quality of clause generation based on the reference plans by comparing them against the reference (expected) clauses.

The BART-based generative model outperforms all the baselines consistently.
The significantly lesser performance of the GPT-2-based approach showcases the merit in conditional over causal generation for this problem.
The difference in performance can also be observed in the Prompt2Clause (causal) and the Top2Clause (conditional) baselines.
Plan2Clause-Retrieval turns out to be reasonably competitive, indicating the effectiveness of the proposed method for keyword planning and the potential advantage due to the similar nature of clauses.
The importance of an ordered plan can be gauged from the poorer performance of the RandKwds2Clause baseline.

Although {\sc ClauseRec} is a pre-existing literature in the generation of legal clauses, it works on a contract-level problem of recommending a new clause to an incomplete contract in contrast to our clause-level problem.
The significant difference in the input and nature of the problem hinders a direct comparison between the two approaches.
However, our work merits in demonstrating extensibility (refer Section \ref{appdx:robust}) over clause topics compared to the previous work which was limited to a selected set of 5 high-frequency topics.

\section{Analysis}

\begin{figure}[h]
    \centering
    \includegraphics[scale=0.35]{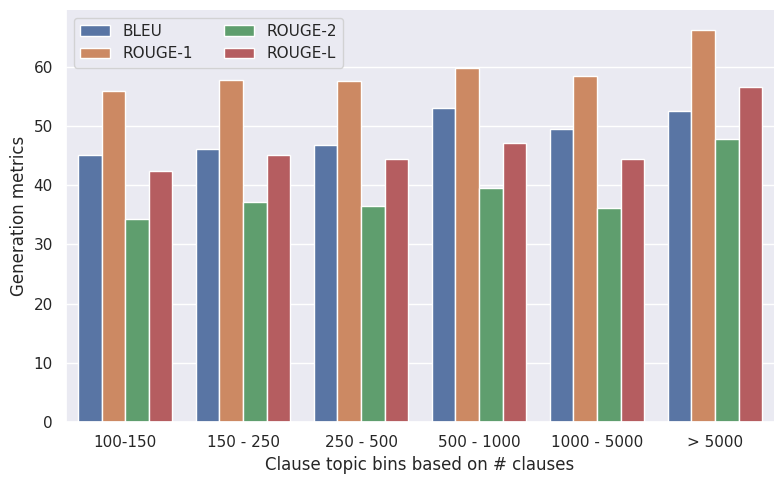}
    \caption{Illustration of the robustness of pipeline to clause topics of various frequencies.}
    \label{fig:robust}
    \vspace{-0.15in}
\end{figure}

\subsection{Robustness across clause topic frequencies}
\label{appdx:robust}

Table \ref{tab:clausetypes} shows statistics of the distribution of clause topics in the dataset for different bins based on their frequencies. 

As can be seen, a significant variation exists in the nature of clause topics: a large number of clause topics with fewer clauses per topic coexist with a small number of clause topics with a very high number of clauses.
We inspected the generation performance across multiple bins of clause frequencies to analyze the difference in generation quality.
As seen in Figure \ref{fig:robust}, there is only a small difference in the performance of the lowest frequency bin compared to the highest.
This is despite the number of clauses under the lowest frequency topics being almost 2 orders of magnitude lesser than the highest frequency ones.
This shows no significant bias towards the clause topics with a very high number of clauses, and the model can handle a diverse range of topics well.
These observations demonstrate the extensibility of our approach to handle newer topics with fewer clauses per topic.

\subsection{Comparison with sequential keyword order}

We also ablate our proposed generic to specific order of keywords against a natural, sequential order for the planning and generation stages.
Much of the existing literature uses a sequential content plan as the conditional prior for generation.
This is natural, since the models typically work by generating the content sequentially based on the keyword information, which intends to plan a `story' in that order.

\subsubsection{Planning}

\begin{figure}[H]
    \centering
    \includegraphics[scale=0.36]{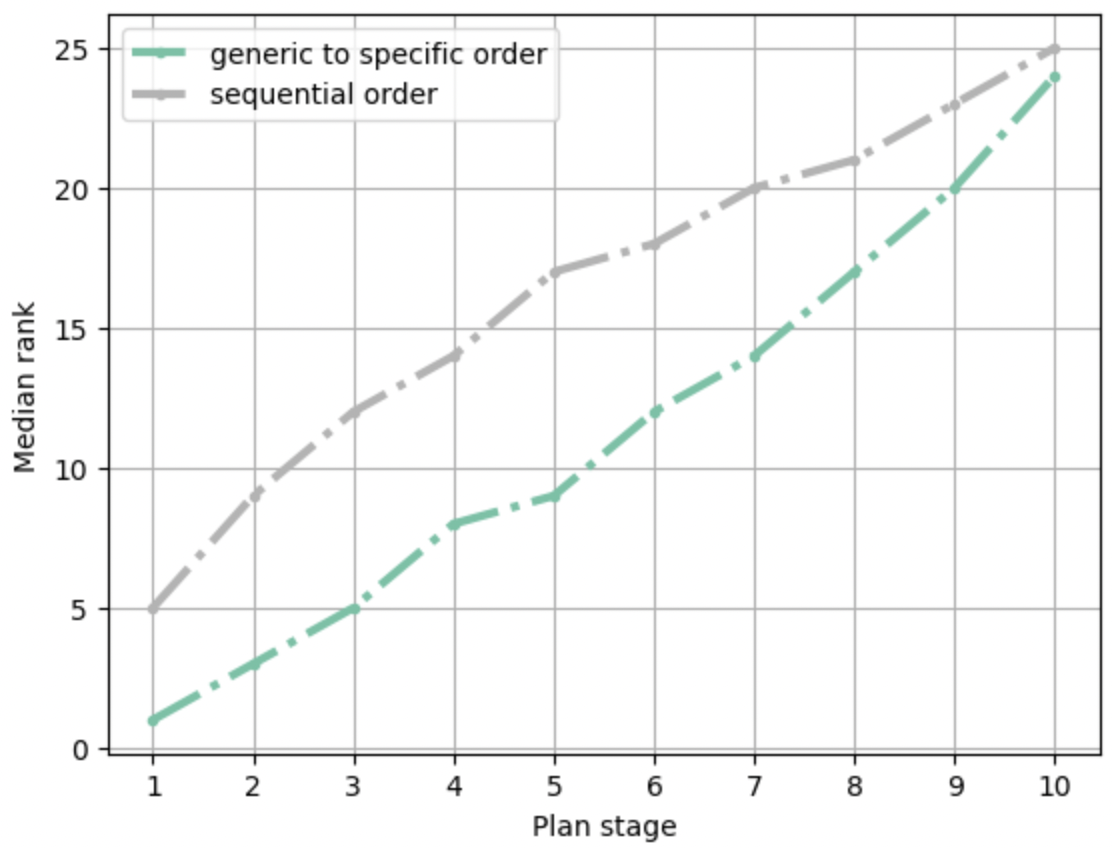}
    \caption{Comparison of stage-wise median ranks given to plans generated by generic to specific ordered keywords versus sequential keywords (the lower the better).}
    \label{fig:planrank-kwd-seq}
\end{figure}

Figure \ref{fig:planrank-kwd-seq} shows the median ranks given to the plans generated by our graph-based approach to each stage of the 10-stage plan.
For generating plans with the sequential keywords approach, the first top 10 keywords from every clause were extracted to prepare the dataset $D^t \: \forall \: t \in T$, following which the same procedure was followed for graph generation and planning.

As can be observed from the figure, the proposed approach guided by topic-level information for keyword extraction performs better than the approach based on keyword extraction at a clause level for the initial stages of the plan.
The lower ranks given in the initial stages of planning highlight the predictive components in both the approaches.
There are only a few possibilities for topic-generic content keywords for the proposed generic to specific approach.
In the sequential approach, the initial lower ranks bring out the predictable nature of initial phrases in a legal clause.
As we move to later stages, the gap between the two approaches decreases as the ranks increase.
This demonstrates the loss of predictability as we move on to increase the no. of plan keywords.

\subsubsection{Generation}

\begin{figure}[H]
    \centering
    \includegraphics[scale=0.55]{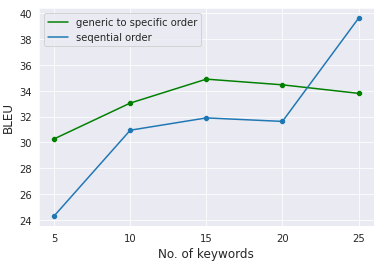}
    \caption{Ablation analysis based on the number of keywords for clause generation \& comparison of the proposed content plan order against the traditional sequential keyword order.}
    \label{fig:gen-kwd-seq}
    \vspace{-0.15in}
\end{figure}

In order to study the impact of the no. of keywords provided for generation, we conducted an ablation analysis by repeating our experiments for clause generation on a BART-base model for 10 epochs each, keeping the rest of the hyperparameters the same.
The no. of keywords was changed from 5 to 25 in step of 5 for conducting these experiments.

To contrast our method with the sequential keyword order, we repeated the ablation experiments for the same number of keywords while considering the content plan as a sequential order of keywords in the clause.
We measured the performance across these studies using BLEU and have illustrated the results in Figure \ref{fig:gen-kwd-seq}.

For our proposed keyword order, we found the generation performance to initially increase with an increase in the number of keywords but later follow a decreasing trend.
As we increase the number of keywords for generation, the number of keywords specific to a clause increases.
This could help initially since the generative model gets more context for generation. However, adding too many clause-specific keywords away from the topic confounds the model by supplying too much information.
Another factor that could play here is the increased noisiness in keywords as we move down the ranked hierarchy, where the keywords keep getting less indicative of the content.

However, when providing the keywords sequentially, the performance followed a continuously increasing trend.
The increase seems natural since the generative model has to only fill in a lesser amount of content in the already provided natural order of keywords as we keep providing more keywords.
So the task of generation for the model keeps getting easier.
However, the sequential order performs markedly poorly against our approach for the initial stages before crossing over to higher scores, as shown in Figure \ref{fig:gen-kwd-seq}.
This shows the merit of our approach for clause generation based on minimal keyword information, helping the system to reach the desired clause quickly - without asking for a more significant number of keywords.
It is important to appreciate the practicality aspect here since we may not expect an end user to keep providing input to the extent that the basis of the motive lying in automated clause generation gets moot.

\subsection{Generation quality}

\begin{table*}[h]
    \centering
    \scalebox{0.85}{
    \begin{tabular}{P{0.1\linewidth} | m{\linewidth}}
        \hline
        Topic & advances \\
        \hdashline
        Plan & advance, amount, section, revolving, rate, accordance, outstanding, interest, principal, subject \\
        \hdashline
        Generated & \small{Subject to section 2.2(b), the principal amount outstanding under the revolving line shall accrue interest at a floating per annum rate equal to the prime rate plus the applicable margin, which interest shall be payable monthly in accordance with the terms of this agreement.} \\
        \hdashline
        Actual & \small{Subject to section 2.3(b), the principal amount of advances outstanding under the revolving line shall accrue interest at a floating per annum rate equal to the prime rate, which interest shall be payable monthly in accordance with section 2.3(d) below.} \\
        \hline
        \multicolumn{2}{c}{} \\
        \hline
        Topic & brokers \\
        \hdashline
        Plan & broker, connection, commission, fee, agreement, finder, transaction, contemplated, brokerage, entitled \\
        \hdashline
        Generated & \small{No broker, finder or investment banker is entitled to any brokerage, finders or other fee or commission in connection with the transactions contemplated by this agreement based upon arrangements made by or on behalf of buyer.} \\
        \hdashline
        Actual & \small{Except for Raymond James Financial, Inc. and Oppenheimer \& Co. Inc., no broker, finder or investment banker is entitled to any brokerage, finder's or other fee or commission in connection with the transactions contemplated by this agreement or any other transaction agreement based upon arrangements made by or on behalf of strand.} \\
        \hline
        \multicolumn{2}{c}{} \\
        \hline
        Topic & data privacy \\
        \hdashline
        Plan & data, company, plan, personal, transfer, participation, administering, consent, purpose, managing \\
        \hdashline
        Generated & \small{The grantee hereby explicitly and unambiguously consents to the collection, use and transfer, in electronic or other form, of his or her personal data as described in this agreement and any other award grant materials (data) by and among, as applicable, the employer, the company and any parent or subsidiary for the exclusive purpose of implementing, administering and managing the grantees participation in the plan.} \\
        \hdashline
        Actual & \small{The grantee hereby voluntarily consents to the collection, use and transfer, in electronic or other form, of the grantees personal data as described in this agreement and any other award grant materials by and among, as applicable, the company and any subsidiary or affiliate for the exclusive purpose of implementing, administering and managing the grantees participation in the plan.} \\
        \hline
        \multicolumn{2}{c}{} \\
        \hline
        Topic & withholdings \\
        \hdashline
        Plan & withholding, tax, applicable, payment, agreement, pursuant, subject, made, income, employment \\
        \hdashline
        Generated & \small{All payments made pursuant to this agreement shall be subject to withholding of applicable income and employment taxes.} \\
        \hdashline
        Actual & \small{All payments made pursuant to this agreement will be subject to withholding of applicable income, employment and excise taxes.} \\
        \hline
        \multicolumn{2}{c}{} \\
        \hline
        Topic & limitation of liability \\
        \hdashline
        Plan & damage, party, consequential, indirect, punitive, incidental, notwithstanding, foregoing, entitled \\
        \hdashline
        Generated & \small{The collateral trustee shall not be liable for any action taken or omitted to be taken by it hereunder or under any other secured debt document, except for its own gross negligence or willful misconduct.} \\
        \hdashline
        Actual & \small{The collateral trustee will not be responsible or liable for any action taken or omitted to be taken by it hereunder or under any other security document, except for its own gross negligence or willful misconduct as determined by a final order of a court of competent jurisdiction.} \\
        \hline
    \end{tabular}
    }
    \caption{Example clauses generated by the best performing BART model given a topic with the corresponding plans and actual (reference) clauses.}
    \label{tab:examples}
    \vspace{-0.15in}
\end{table*}

We show a few examples of clauses generated from reference plans in Table \ref{tab:examples}.
The generation quality observed, backed by the aggregated quantitative results showcases the clause generation model's efficacy in generating appropriate clauses from their corresponding plans.

The generated clauses succeed in conveying the same intent as the actual clause while also being strikingly similar in their lexical content.
The clauses seem to capture the nuances in legal writing very well, and also change based on the topic and context of the content being generated.
The generated clauses naturally fail to add any entity-specific information (such as the clause shown under the topic \textit{brokers}) since the approach does not account for taking in these inputs from the user.
Future work can explore incorporating such information from the user to generate entity-specific clauses.
Considering contract-specific reference information (such as ``section 2.2(b)'' in the generated clause under the topic \textit{advances} as opposed to ``section 2.3(b)'') can be yet another scope for future work.
Many of the differences between the actual and generated clause content involved phrases implying similar intent, such as ``explicitly and unabiguously'' versus ``voluntarily'' for the example under \textit{data privacy}.
Most of the shorter length clauses showed good lexical overlap with the actual (e.g. \textit{withholdings}) with hallucination observed in some content (e.g. \textit{limitation of liability}).

Besides entity and contract-specific information, future work can also handle allowing phrase-level control in clauses, where the user can ask for explicit phrases to be included, or not to be included in the clause along with specification of a few custom keywords for reference.
The challenge here would lie in detecting the appropriateness of position for placement of that phrase within the clause.

\subsection{Controllable, iterative plan-to-clause workflow}

We demonstrate the controllability in clause generation with an example flow of iterative clause customization in Appendix \ref{appdx:iterative-flow}.
We found the generated clauses to suitably vary content based on simple addition and removal of necessary keywords which can encourage approaches for developing efficient tools in legal clause drafting.

Consider an end user to be acquainted with drafting legal contracts - for instance lawyers.
An iterative flow involving content planning followed by clause generation allows the user to keep deleting and adding keywords to the plan for driving down towards a desired state of the clause.
The idea is to allow an end-user to generate a legal clause by the specification of minimal information such that the final generated state of the clause can be used with minimal edits necessary.
A keyword-based information control facilitates this simplicity compared to control based on latent space representation.

An interesting problem we believe future work could look at is ensuring only necessary phrasal changes are made between two successive stages of clause generation in the iterative pipeline shown in Figure \ref{fig:iterative-flow}, thus making the control more precise.
For instance, the addition of the keyword ``law'' to the plan in the third stage of generation makes changes to the clause like changing ``governmental authority'' to ``arbitrator'', increasing the verbosity of the clause and a slight change of meaning w.r.t. company's shares of common stock.
Explainability and more nuanced control in this process would make clause generation more precise.

\section{Conclusion}
We propose a plan-based approach for generating legal clauses inspired by content planning techniques in story generation.
The pipeline involves customizable content plan generation based on the clause topic and optional control keywords using a simple, lightweight graph followed by clause generation.
The content plan represents its corresponding clause as an ordered list of generic to specific keywords.
Our approach achieves promising results for clause generation across the broad range of clause topics in the dataset, indicating the extensibility of our approach.
We also show the merit of our proposed order in generating clauses with lesser keyword information.
While we discuss a use case for controllability in clause generation possible through our pipeline, the generation of clause content shows substantial changes for minor changes in the plan.
Future work can look at increasing the preciseness of control involved by changing only the content of a clause as necessitated by a change in the input plan.
The customization of clause content can be further drilled down to inclusion of entity-specific and contract-specific information.

\section{Limitations}

While we evaluate the generation of clauses by using regular generation-based metrics (BLEU \& ROUGE), establishing results based on human evaluations would have provided substantial qualitative backing for the empirically strong results.
However, the understanding and evaluation of clauses would require strong domain knowledge in legal clauses, and any evaluation from a layperson would not help in gauging the quality.
Due to the practical difficulties in involving domain-specific experts to evaluate a substantial number of clauses to make a judgment, we relied on the quantitative metrics and some qualitative analysis performed randomly on a select set of clauses.

Controllable content generation has been popularly demonstrated by the CTRL \cite{CTRL} architecture that shows fine capabilities in controlling the content based on the specification of control keywords appended before the prompt for a generation.
Although it would have been interesting to study the performance of this model fine-tuned for clause generation, we were limited by sufficient computational resources to carry out the experiment on this model, and on the larger variants of models (BART, GPT-2) we currently have used.

\bibliography{custom}
\bibliographystyle{acl_natbib}

\appendix

\section{Frequency distribution of clause topics in the dataset}
\label{appdx:topic-distri}

Aggregated statistics of bins of clause topics based on their frequencies are given in Table \ref{tab:clausetypes}.
By the frequency of a clause topic, we mean the number of clauses under that topic.

\begin{table}[h]
\scriptsize
    \centering
    \begin{tabular}{|c|c|c|c|c|}
        \hline
        range & \# topics & mean & std. & median \\
        \hline
        100-150 & 309 & 120.5 & 14.2 & 118.0 \\
        150-250 & 233 & 192.1 & 29.2 & 186.0 \\
        250-500 & 230 & 344.3 & 65.5 & 332.0 \\
        500-1k & 102 & 660.7 & 119.2 & 643.5\\
        1k-5k & 62 & 1901.9 & 912.2 & 1483.5 \\
        >5k & 5 & 8205.2 & 1304.7 & 7853.0 \\
        \hdashline
        Overall & 939 & 412.4 & 764.5 & 210.0 \\
        \hline
    \end{tabular}
    \caption{Distribution of clause topics in the dataset w.r.t. the number of clauses under each topic. The values of mean, median, and std indicate the corresponding values of the number of clauses under a topic in that bin range. The topic \textit{governing laws} had the highest number of clauses at 10,636.}
    \label{tab:clausetypes}
\end{table}

\section{Iterative planning and generation: Example use case}

\label{appdx:iterative-flow}
\begin{figure*}[t]
    \vspace{-12cm}
    \centering
    \includegraphics[scale=0.5]{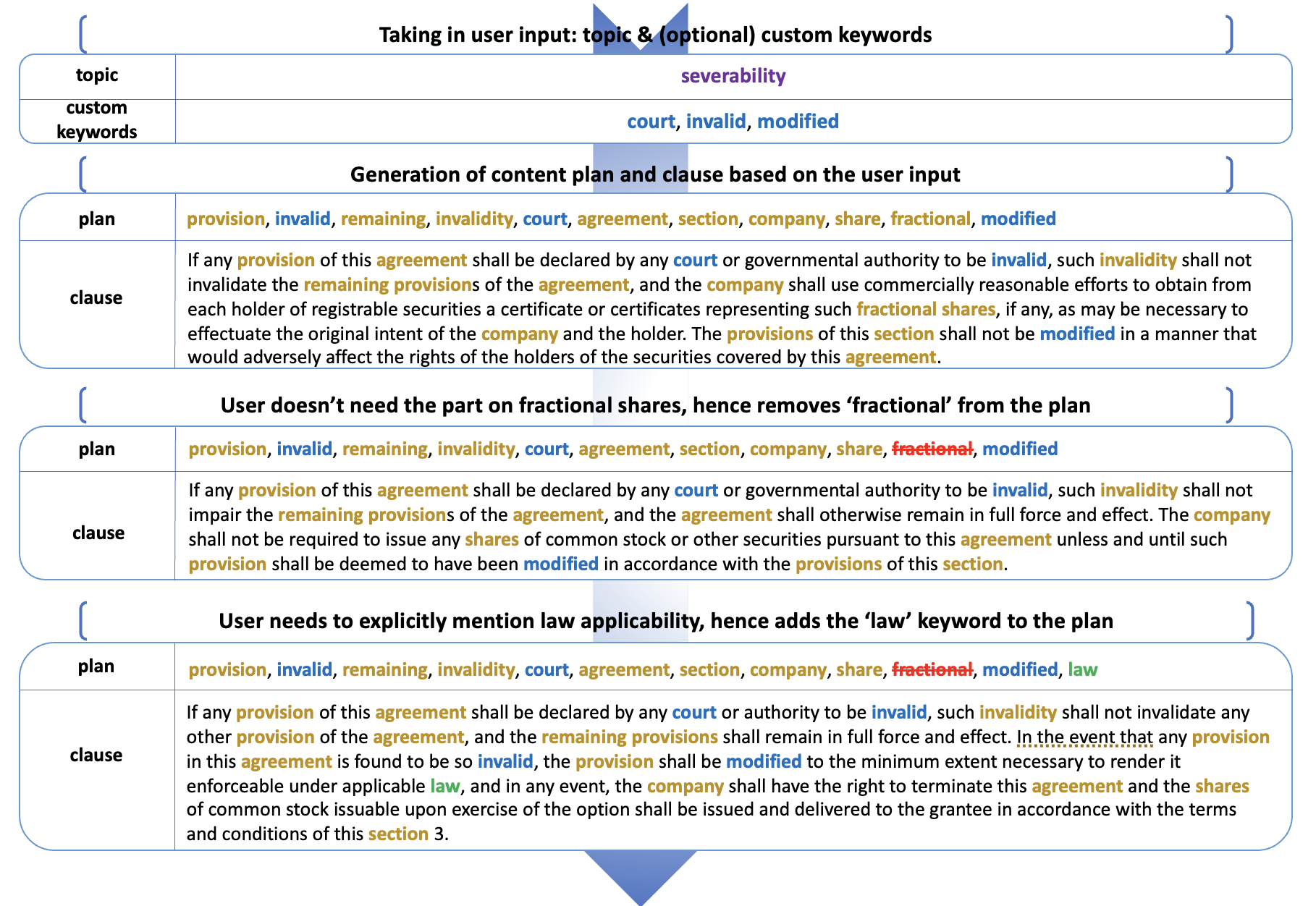}
    \caption{Example flow where an end user generates a clause for the clause topic \textit{severability} by specifying a few additional keywords at the start. In subsequent stages, the user removes and adds keywords from the generated plan to directly control the clause content.}
    \label{fig:iterative-flow}
\end{figure*}

Figure \ref{fig:iterative-flow} shows an example use case of the customized plan to clause generation. 

\end{document}